\documentclass[runningheads]{llncs}

\usepackage[mobile]{eccv}

\usepackage{eccvabbrv}

\usepackage{graphicx}
\usepackage{booktabs}
\usepackage{multicol}
\usepackage{multirow}
\usepackage[table]{xcolor}
\usepackage{enumitem}
\usepackage[ruled,vlined]{algorithm2e}
\usepackage{algpseudocode} 

\usepackage[accsupp]{axessibility}  

\let\titleold\title
\renewcommand{\title}[1]{\titleold{#1}\newcommand{\thetitle}{#1}}
\newcommand{\maketitlesupplementary}{
    \newpage
    \begin{center}
        \Large
        \textbf{\thetitle} \\
        \vspace{0.5em}
        \large  Supplementary Material  \\
        \vspace{1.5em}
    \end{center}
}

\usepackage{hyperref}

\usepackage{orcidlink}

\begin{document}

\title{NanoGS: Training-Free\\ Gaussian Splat Simplification} 

\titlerunning{NanoGS}

\author{Butian Xiong
\inst{1}\thanks{Co-first authors, equal technical contribution.} \and
Rong Liu
\inst{1}\protect\footnotemark[1]
\and
Tiantian Zhou\inst{1}
\and\\
Meida Chen\inst{1}
\and
Zhiwen Fan\inst{2}
\and
Andrew Feng\inst{1}}

\authorrunning{Xiong et al.}

\institute{USC Institute for Creative Technologies \and
Texas A\&M University}

\maketitle
\begin{figure}
    \centering
    \includegraphics[width=0.99\linewidth]{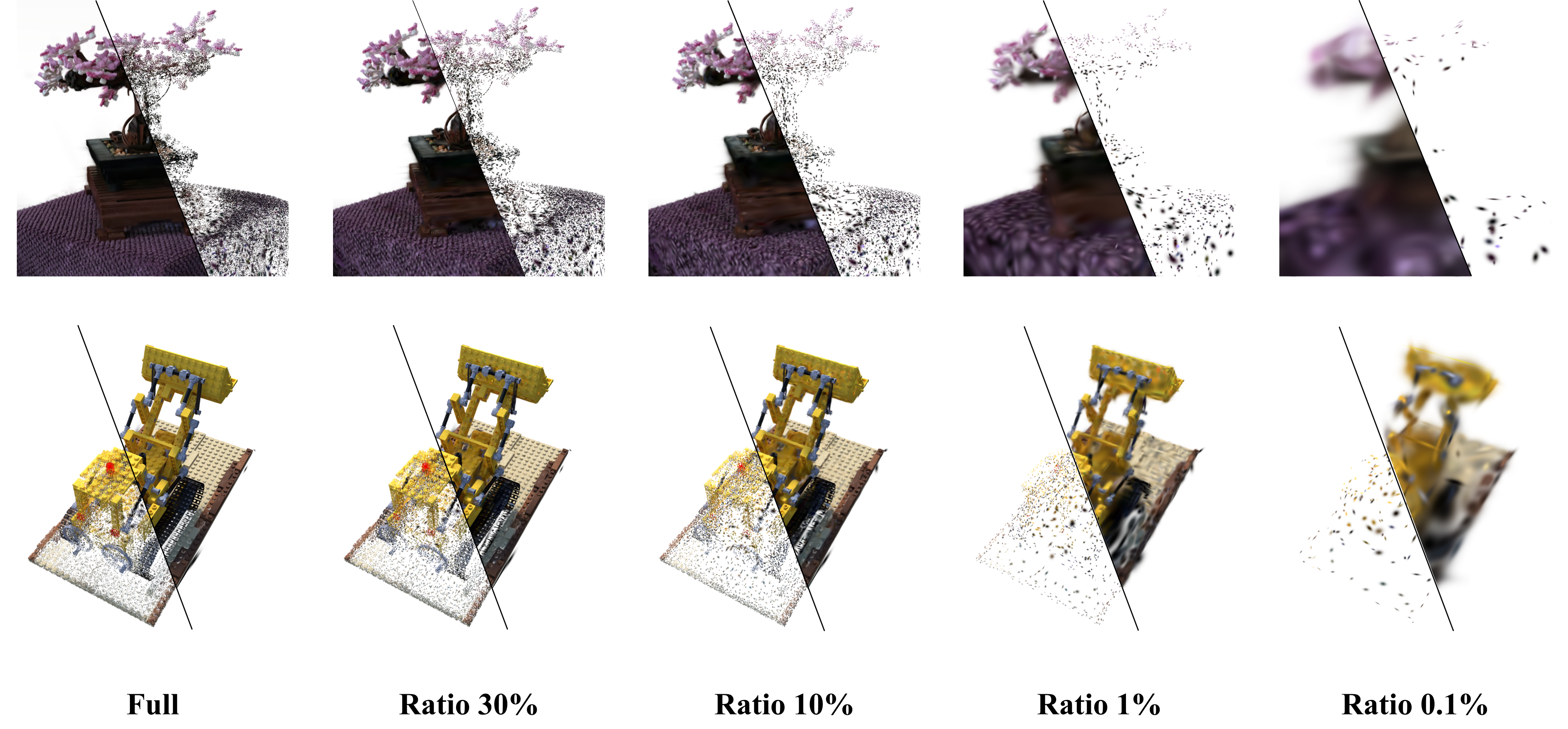}
    \caption{\textbf{NanoGS} reduces Gaussian Splat primitive count from the full model to increasingly compact representations. \textbf{NanoGS} achieves substantial compaction ratio while preserving visual fidelity and geometric structure without GPU-intensive optimization or calibrated images.}
    \label{fig:teaser}
\end{figure}

\begin{abstract}
3D Gaussian Splat (3DGS) enables high-fidelity, real-time novel view synthesis by representing scenes with large sets of anisotropic primitives, but often requires millions of Splats, incurring significant storage and transmission costs. Most existing compression methods rely on GPU-intensive post-training optimization with calibrated images, limiting practical deployment.
We introduce \textbf{NanoGS}, a training-free and lightweight framework for Gaussian Splat simplification. Instead of relying on image-based rendering supervision, NanoGS formulates simplification as local pairwise merging over a sparse spatial graph. The method approximates a pair of Gaussians with a single primitive using mass preserved moment matching and evaluates merge quality through a principled merge cost between the original mixture and its approximation. By restricting merge candidates to local neighborhoods and selecting compatible pairs efficiently, NanoGS produces compact Gaussian representations while preserving scene structure and appearance.
NanoGS operates directly on existing Gaussian Splat models, runs efficiently on CPU, and preserves the standard 3DGS parameterization, enabling seamless integration with existing rendering pipelines. Experiments demonstrate that NanoGS substantially reduces primitive count while maintaining high rendering fidelity, providing an efficient and practical solution for Gaussian Splat simplification.
Our project website is available at \href{https://saliteta.github.io/NanoGS/}{https://saliteta.github.io/NanoGS/}
.
\end{abstract}
\section{Introduction}
\label{sec:intro}

Real-time radiance field rendering has been fundamentally reshaped by 3D Gaussian Splatting (3DGS) \cite{3dgs}. By representing scenes as collections of anisotropic 3D Gaussians optimized with adaptive density control, 3DGS achieves photorealistic novel view synthesis while sustaining real-time 1080p rendering performance. Its explicit, rasterization-compatible formulation bridges Neural Radiance Field (NeRF) \cite{mildenhall2021nerf} and classical graphics pipelines, eliminating costly volumetric ray sampling and enabling practical deployment. As a result, Splat-based representations have rapidly become a dominant alternative to NeRF-style models \cite{mildenhall2021nerf,barron2021mipnerf,barron2023zipnerfantialiasedgridbasedneural,M_ller_2022_instantngp}, supporting applications in immersive exploration \cite{jiang2024vrgsphysicaldynamicsawareinteractive, franke2024vrSplattingfoveatedradiancefield}, dynamic scene reconstruction \cite{wu20244dgaussianSplattingrealtime,yang2023deformable3dgaussianshighfidelity,yang2024realtimephotorealisticdynamicscene}, 3D content creation \cite{yi2023gaussiandreamer, tang2023dreamgaussian}, semantic distillation \cite{qin2024langSplat,zhou2024feature3dgssupercharging3d,xiong2025Splatfeaturesolver}, and cinematic production \cite{wang2024cinematic}.

Despite this success, a fundamental bottleneck persists: \textbf{scalability}, as high-fidelity reconstructions routinely require millions of primitives to capture fine geometric structures and complex view-dependent appearance. In practice, Splat models frequently occupy hundreds of megabytes or even gigabytes, while large primitive sets increase sorting and rasterization overhead during rendering. This structural scaling limits rendering speed, efficient transmission, edge-device deployment, and large-scale distribution of Splat-based content.

Existing methods mitigate this issue through optimization of the parameter dimensionality, quantization of attribute precision, and reduction of primitive counts \cite{3DGSzip2025, compressionSurvey}. A large body of work focuses on parameter- and bit-level compression through vector quantization, entropy coding, adaptive spherical-harmonic reduction, structured parameter layouts, and progressive encoding schemes~\cite{Reduced3DGS, compact, Compressed3DGS, eagles, lightGaussian}. Other methods integrate pruning or sparsification into retraining pipelines, leveraging importance metrics, regularization, distillation, or optimal-transport-based merging to reduce the number of Splats~\cite{zhang2025gaussianspa, lee2025optimizedminimal3dgaussian, GHAP}. However, these approaches often rely on GPU-intensive optimization and calibrated image supervision. This assumption does not hold in many practical scenarios where Splat models are produced through alternative pipelines, such as 3DGS generation \cite{tang2023dreamgaussian, yi2023gaussiandreamer}, edition \cite{superSplat, guedon2024frosting}, or conversions from meshes or point clouds \cite{zhou2024gaussianpainterpaintingpointcloud, mesh2Splat}, where guided images are not be available. In addition, several compression schemes modify the underlying Gaussian Splat parameterization by introducing codec-specific layouts or structured representations, breaking compatibility with standard Splat rendering pipelines and preventing the compressed models from being directly reused. Consequently, a training-free, lightweight, and representation-preserving simplification framework that operates \emph{post hoc} on existing 3DGS models remains largely unexplored.

A natural objection is that one could simply re-optimize a smaller model under a tighter primitive budget rather than compact an existing one post hoc. This is viable for object-level or synthetic scenes where the original capture pipeline is on hand, but breaks down under three conditions common at deployment time:

\begin{itemize}
    \item \textbf{Disjoint create/consume hardware.} Reconstruction relies on CUDA-enabled clusters, whereas end users render on smartphones, AR headsets, and WebGL clients that lack the VRAM and stack for an on-device optimization loop, where even efficient on-device training and rendering remain open~\cite{pocketgs, du2026mobile-gs}. Under a \emph{create once, publish everywhere} model~\cite{api, khronos_3dcommerce_guidelines}, and with transmission formats carrying only compiled geometry rather than source imagery~\cite{khronos_gltf_gs}, a single asset must serve all devices.
    \item \textbf{Missing supervision.} The dense multi-view supervision that optimization-based compaction needs is often unavailable: source images may be permanently inaccessible (scanners and drone surveys gatekeep raw imagery), and diffusion-driven scene generation produces no ground-truth views~\cite{luciddreamer2023, gaussiancity2025}. Even feed-forward single-image pipelines~\cite{sharp2025} supply one view and pose but not the coverage needed to constrain a photometric optimization, leaving re-optimization under-determined.
    \item \textbf{Poor fit to on-the-fly decimation.} A creator rarely knows a client's memory ceiling in advance, and re-running an hours-long optimization per budget is impractical. Resampling new views does not help, as view planning is isomorphic to set cover and thus NP-hard~\cite{scott2003viewplanning}, and heuristic sampling introduces occlusion and floater artifacts requiring dedicated repair~\cite{gsfix3d2025}.
\end{itemize}

In this work, we introduce \textbf{NanoGS}, a training-free and lightweight \emph{framework} for Gaussian Splat simplification. Rather than a single merging algorithm, NanoGS decouples simplification into three interchangeable stages—a \emph{candidate topology} that proposes which splats may merge, a \emph{cost objective} that scores each candidate, and a \emph{splat operator} that fuses a selected pair—so that each can be swapped without redesigning the others. We deliberately instantiate them with lightweight choices (a sparse $k$-nearest-neighbor graph, a Monte~Carlo I-divergence merge cost, and mass-preserving moment matching) to show that strong extreme-compaction quality follows from the decoupled structure rather than from operator complexity. Concretely, NanoGS first prunes low-opacity splats, then iteratively builds the KNN graph, scores candidate edges by how well a two-Gaussian mixture is approximated by a single Gaussian via moment matching, and greedily collapses low-cost disjoint pairs into mass-preserving Gaussians until the target ratio is reached.
Experiments show that NanoGS substantially reduces primitive count while maintaining high visual fidelity compared to state-of-the-art (SOTA) compression methods. 
Specifically, across four standard benchmarks (NeRF-Synthetic, Mip-NeRF 360, Tanks\&Temples, Deep Blending; 21 scenes total) and three compaction budgets $\rho\in\{0.1,0.01,0.001\}$, NanoGS achieves the best PSNR at every budget and improves over the compared methods by \textbf{+2.40 dB}, \textbf{+4.84 dB}, and \textbf{+5.46 dB} on average, respectively.
Importantly, NanoGS operates directly on existing Gaussian Splat models without requiring training images or optimization, preserving the standard representation and remaining fully compatible with existing rendering pipelines. Unlike level-of-detail schemes that couple spatial structure to merging logic for transient rendering\cite{kerbl2024hierarchical}, NanoGS keeps the two separable and targets permanent asset compaction, a distinction we revisit in Sec.~\ref{sec:discussion} once the stages are formalized.

In summary, our contributions are as follows:
 
\begin{enumerate}
    \item \textbf{A modular simplification framework} that casts training-free, post-hoc simplification as a decoupled pipeline of three interchangeable stages candidate topology, cost objective, and splat operator. This structure, rather than any single operator, drives quality under aggressive compaction and separates our permanent-compaction setting from rendering-time level-of-detail schemes.
    \item \textbf{Efficient graph pairing} (topology stage): a sparse spatial neighbor graph with greedy disjoint-pair selection for scalable simplification.
    \item \textbf{An I-divergence merge cost} (objective stage): scoring edges by the I-divergence between the two-splat mixture and its moment-matched Gaussian, plus an appearance discrepancy.
    \item \textbf{Mass-Preserved Moment Matching} (operator stage): a principled fusion of two splats into one Gaussian preserving geometric and appearance consistency. Stages~2--4 are deliberately lightweight instantiations chosen to isolate the framework's contribution.
    \item \textbf{State-of-the-art extreme compaction} with accessible, CPU-only, representation preserving deployment, attaining the best PSNR at every tested budget and the largest margins in the most aggressive regimes.
\end{enumerate}
\section{Related Works}
\label{sec:related_works}

Recent work has explored improving the efficiency of 3D Gaussian Splatting (3DGS) models along two complementary directions: \emph{compression} and \emph{compaction}~\cite{3DGSzip2025, compressionSurvey}. 
\emph{Compression} methods aim to reduce the storage footprint and transmission bandwidth of 3DGS assets through efficient parameter representations and attribute quantization. 
In contrast, \emph{compaction} methods focus on reducing the number of Gaussian primitives while preserving rendering quality, thereby lowering both memory consumption and rasterization cost during rendering. 
These two directions address different aspects of efficiency and are largely orthogonal: compression minimizes the bit-level storage of its parameters, while compaction simplifies the structural complexity of the representation.

\subsection{3DGS Compression: Bit-Level Storage Reduction}

While 3D Gaussian Splatting (3DGS) enables real-time photorealistic rendering, early models often produce serialized assets ranging from hundreds of megabytes to several gigabytes, creating a major bottleneck for storage, transmission, and web-based deployment~\cite{compressionSurvey, Compressed3DGS, HAC}. To address this issue, a large body of work focuses on \emph{bit-level compression} of Gaussian attributes. These methods typically quantize and entropy-code per-Gaussian parameters such as position, covariance, color coefficients, and opacity, often combined with learned or engineered context models and auxiliary coding structures, including anchor-based predictors and hierarchical entropy models~\cite{HAC, HAC++, CONTEXTGS, HEMGS, Compressed3DGS}.

Representative approaches include Compressed3DGS~\cite{Compressed3DGS}, which employs sensitivity-aware clustering and quantization-aware fine-tuning to achieve up to $31\times$ compression while maintaining rendering quality, and provides a WebGPU-based renderer for efficient browser visualization. Other methods improve storage efficiency by reducing the \emph{per-primitive parameter payload} through compact parameterizations or shared representations, such as reduced bases, codebooks, or alternative primitive formulations, sometimes combined with auxiliary pruning strategies for additional memory savings~\cite{Reduced3DGS, CompactSLAM3DGS, DBS, UBS}.
Note that our method belongs to the compaction category, reducing the number of Gaussian primitives while preserving the standard 3DGS representation. As a result, it remains fully compatible with existing compression techniques, which can be applied on top of the compacted model to further reduce the overall storage footprint.

While these approaches significantly reduce the storage cost of 3DGS assets, many rely on access to the calibrated training images to optimize compression modules. This assumption does not always hold in practice, particularly when Gaussian Splat models are generated, edited, or converted through alternative pipelines where training images are not required. Moreover, several methods introduce codec-specific parameterizations or auxiliary structures that may limit direct compatibility with standard 3DGS rendering pipelines.

\subsection{3DGS Compaction: Primitive-Level Count Reduction}

While sharing the goal of reducing the number of Gaussian primitives, existing compaction approaches differ substantially in their operational requirements and optimization strategies.

\textbf{Training-based compaction.}
A line of work incorporates primitive reduction directly into the 3DGS training pipeline so that the model converges to fewer splats or a constrained primitive budget~\cite{LP-3DGS, compact, taming3DGS}. 
These methods typically introduce additional mechanisms such as learned masking, compact parameterizations, or resource-aware training schedules. 
For example, Compact3DGS~\cite{compact} integrates masking and compact representations into the optimization process, while Taming 3DGS~\cite{taming3DGS} explicitly targets high-quality reconstruction under limited resource budgets by controlling training-time allocation. 
While effective when retraining is possible, such approaches require access to the original training pipeline and therefore cannot serve as drop-in simplification tools for arbitrary pre-trained 3DGS assets.

\textbf{Image-guided pruning.}
Another dominant strategy estimates the importance of each Gaussian by measuring its contribution to rendered pixels in the training views. 
These methods typically require training images, camera poses, and differentiable rendering signals during the compaction process. 
Representative examples include LightGaussian and PUP 3D-GS~\cite{lightGaussian, PUP3DGS}, which compute per-splat importance scores based on reconstruction sensitivity and employ prune-and-recover or prune-and-refine stages to maintain rendering quality under high pruning ratios. 
Although effective, such pipelines depend on rendering supervision and access to the training images, limiting their applicability when only the trained splat model is available. 
In contrast, our approach is \emph{model-only} and operates directly on the Gaussian set without requiring training images or camera parameters.

\textbf{Model-only primitive reduction.}
More closely related to our formulation are classical approaches for Gaussian mixture reduction and clustering based on distributional similarity measures, including Bregman divergences and information-theoretic co-clustering~\cite{bregman-divergence, OptimalityBregman, Bregman-Gaussian, InfoTheoryCoClsutering}. 
These techniques provide principled tools for approximating a mixture of Gaussians with a smaller set while preserving key statistical properties. 
GHAP~\cite{GHAP} is the most closely related work in the 3DGS literature: it formulates the geometry of a splat model as an unnormalized Gaussian mixture and performs blockwise reduction using an optimal-transport-based objective, followed by an appearance refinement stage. 
While effective, it relies on subsequent optimization to restore appearance quality. 

Also related, though more narrowly, is Hierarchical 3DGS (H3DGS)~\cite{kerbl2024hierarchical}, which is comparable to our method specifically at the \emph{merge operator}: it fuses primitives into a bounding-volume hierarchy (BVH) used for initial downsampling and level-of-detail rendering. That our operators resemble each other is unsurprising, since moment matching is the natural way to approximate a two-Gaussian mixture by one Gaussian; any principled pairwise merge arrives at essentially the same operation. The methods differ elsewhere. H3DGS is an optimization-based pipeline that retrains the hierarchy, with the BVH chosen a priori as a downsampling and LOD structure, so it exposes no standalone merge cost or candidate-selection criterion—these are absorbed into training. Our framework is instead training-free and makes \emph{which} pairs merge and \emph{by what cost} the primary objects, leaving the operator interchangeable. The hierarchy we obtain is thus an emergent byproduct of cost-driven merging rather than a prescribed structure, and the topology can be swapped without touching the rest of the pipeline (Sec.~\ref{sec:discussion}).

In contrast, our method focuses on lightweight \emph{post-hoc} simplification of trained splat sets. 
Rather than solving a transport problem, we employ an efficient local merging strategy that approximates pairs of Gaussians via moment matching, enabling training-free compaction that operates directly on existing models while preserving geometric and appearance consistency.

\section{Methods}
\begin{figure*}[t]
    \centering
    \includegraphics[width=\linewidth]{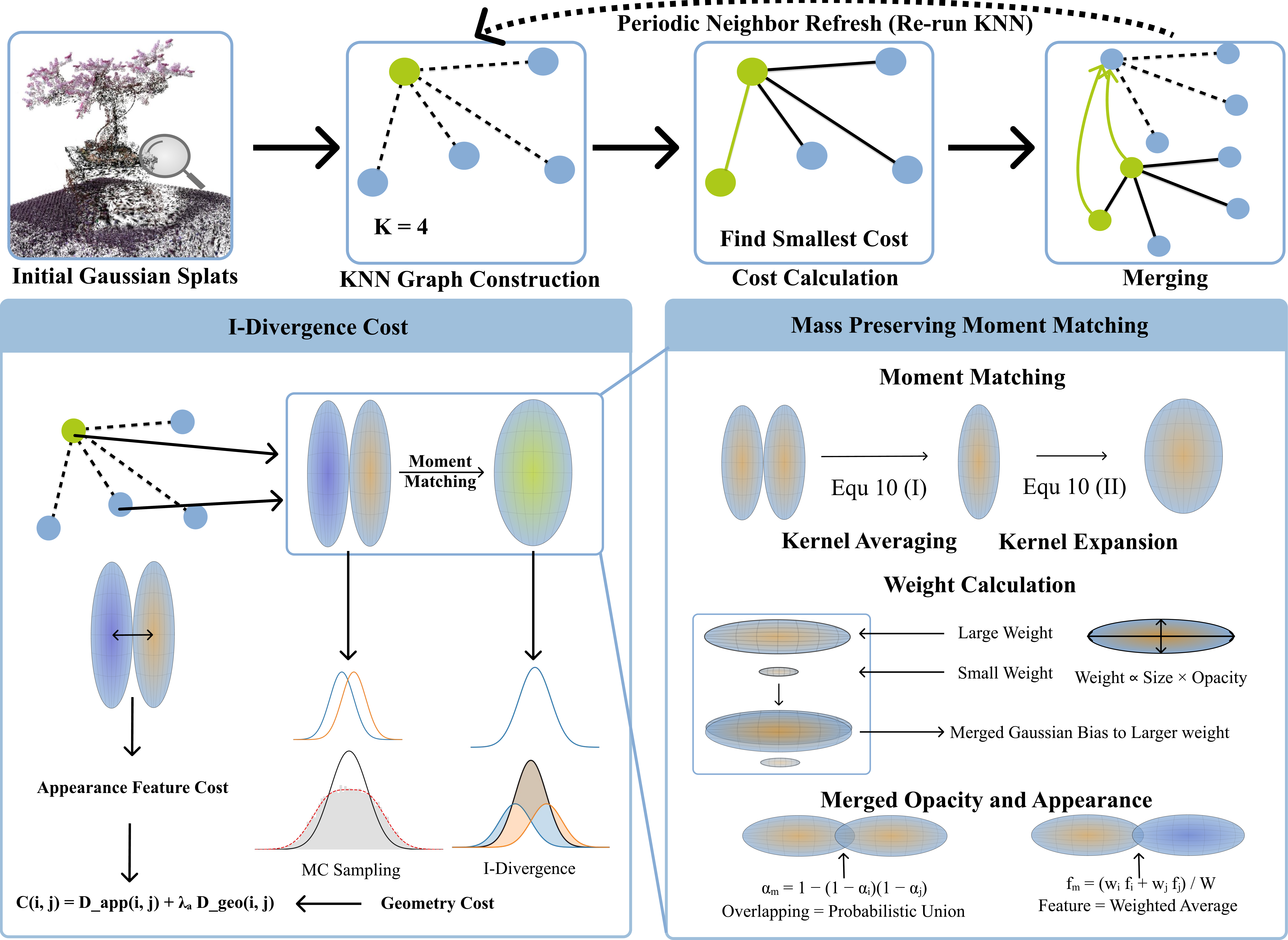}
    \caption{\textbf{NanoGS pipeline overview.} \textit{(Top)} Starting from an initial set of Gaussian splats, NanoGS constructs a sparse $k$-nearest-neighbor graph, evaluates a merge cost for each candidate edge, and collapses the lowest-cost disjoint pairs progressively. \textit{(Bottom-left)} The merge cost combines an appearance term and a geometry term. The geometry cost measures the I-divergence between the original two-splat mixture and its single-Gaussian approximation. \textit{(Bottom-right)} The Mass Preserved Moment Matching(MPMM) merge operator fuses a selected pair by computing mass-weighted moments, so the merged Gaussian is biased toward the larger primitive. Opacity is aggregated as a probabilistic union, and appearance features are blended as a weighted average.}
    \label{fig:pipeline}
    \vspace{-4mm}
\end{figure*}
\label{sec:method}
In this section, we describe the proposed splat simplification framework built around a progressive graph-based merging pipeline. We first organize the unstructured splat set into a sparse $k$-nearest-neighbor (KNN) merge graph. The key idea is to turn an unstructured set of $N$ splats
into a sparse \emph{merge graph} and then repeatedly collapse low-cost edges to construct
a coarse-to-fine hierarchy of splat representations. Our pipeline comprises four modules:
\textbf{(i) sparse merge-graph construction} (Sec.~\ref{sec:graph}),
\textbf{(ii) merge-cost evaluation} (Sec.~\ref{sec:cost}),
\textbf{(iii) a merge operator} that fuses a selected pair into one splat (Sec.~\ref{sec:merge}),
and \textbf{(iv) progressive batched edge collapses} with periodic neighborhood refresh
to reach any target compression level (Sec.~\ref{sec:progmerge}). 

\subsection{Sparse Merge Graph Construction}
\label{sec:graph}

Our merge objective assigns a cost to any pair of splats, but evaluating all
$\binom{N}{2}$ pairs is prohibitive for large scenes. We therefore construct a sparse
\emph{merge graph} $G=(V,E)$ that restricts candidates to spatially local neighbors,
reducing candidate evaluation from $\mathcal{O}(N^2)$ to $\mathcal{O}(kN)$. Figure~\ref{fig:pipeline} (top, second panel) illustrates this step: each node connects to its $k$ nearest
neighbors (dashed edges), forming the candidate set that we will score and prune.

\textbf{KNN neighborhood graph.}
Let each current splat $i\in V$ have center $\mu_i\in\mathbb{R}^3$.
For each $i$, we query its $k$ nearest neighbors in 3D and add undirected edges
$(i,j)$ to $E$. This yields $|V|=N$ and $|E|=\mathcal{O}(kN)$, which keeps candidate merges
local while making edge scoring tractable.

\textbf{What the graph represents.}
Each node corresponds to a (possibly already merged) splat, and each edge is a
\emph{candidate merge}. The rest of the method operates on this graph: we score edges,
pick a set of disjoint edges, merge them in parallel, and update the graph.

\subsection{Merge Cost}
\label{sec:cost}
Intuitively (Fig.~\ref{fig:pipeline}, bottom-left), $\mathcal{D}_{\mathrm{merge}}$ is small when the two
Gaussians overlap enough that a single Gaussian can cover them with little mass mismatch, and large when the
pair is multi-modal (well-separated means), forcing the approximation to either miss probability mass or
over-inflate.
We define the cost of merging two splat sets $i$ and $j$ as a combination of
a \textbf{geometric merge distortion} and an \textbf{appearance discrepancy}:
\begin{equation}
\mathcal{C}(i,j)
=
\mathcal{D}_{\mathrm{geo}}(i,j)
+
\mathcal{D}_{\mathrm{app}}(i,j),
\qquad
\label{Equ:cost}
\end{equation}

\textbf{Geometry discrepancy = merge distortion.}
For a candidate pair $(i,j)$, the geometric term measures the distortion incurred
when replacing the two-splat mixture by a single merged splat:
\begin{equation}
\mathcal{D}_{\mathrm{geo}}(i,j)
\;\doteq\;
\mathcal{D}_{\mathrm{merge}}(i,j\!\to\!m),
\label{eq:Dgeo_main}
\end{equation}
where $m$ denotes the merged splat produced by our merge operator
(Sec.~\ref{sec:merge}).

\textbf{Merge distortion (mixture$\to$Gaussian).}
We regard each splat as an \emph{unnormalized} Gaussian mass density
\begin{equation}
p_i(x)\;\doteq\; w_i\,\mathcal{N}(x;\mu_i,\Sigma_i),
\qquad
w_i \;\doteq\; (2\pi)^{3/2}\,\alpha_i\!\prod_k s_{i,k},
\qquad i\in\mathcal{S},
\label{eq:unnormalized_splats}
\end{equation}
where $\mathcal{S}$ denotes the current splat set and $\alpha_i\in[0,1]$ is opacity.
In Monte Carlo evaluation we draw $S$ samples per candidate pair, with $S$ a small
fixed constant.

Let $p_{ij}(x)=p_i(x)+p_j(x)$ denote the unnormalized mixture, and define the
\emph{normalized} mixture distribution
\begin{equation}
\tilde p_{ij}(x)\doteq \frac{p_{ij}(x)}{W}
=
\pi\,\mathcal{N}(x;\mu_i,\Sigma_i) + (1-\pi)\,\mathcal{N}(x;\mu_j,\Sigma_j),
\qquad
\pi \doteq \frac{w_i}{W}.
\label{eq:norm_mixture}
\end{equation}
We approximate $\tilde p_{ij}$ by the merged Gaussian
$q_m(x)=\mathcal{N}(x;\mu_m,\Sigma_m)$ produced by our VM moment matching, and define
the merge distortion as the I-divergence
\begin{equation}
\mathcal{D}_{\mathrm{merge}}(i,j\!\to\!m)
\;\doteq\;
\mathrm{I}\!\left(\tilde p_{ij}\,\middle\|\, q_m\right)
=
\int \tilde p_{ij}(x)\log\frac{\tilde p_{ij}(x)}{q_m(x)}\,dx.
\label{eq:Dmerge_main}
\end{equation}
Because the mixture term $\log(\pi\mathcal{N}_i+(1-\pi)\mathcal{N}_j)$ has no closed form,
we estimate~\eqref{eq:Dmerge_main} with a small fixed number of Monte Carlo samples per
candidate pair.

\textbf{Appearance difference.}
Each splat also carries an appearance feature vector
$f_i \in \mathbb{R}^F$ (e.g., spherical harmonics coefficients).
We define the appearance discrepancy as the squared $\ell_2$ distance:
\begin{equation}
\mathcal{D}_{\mathrm{app}}(i,j)
\;\doteq\;
\left\| f_i - f_j \right\|_2^2 .
\end{equation}

\subsection{Merging Operation}
\label{sec:merge}

\textbf{Mass-preserving moment matching (MPMM).}
Given a candidate pair of splats $(i,j)$, we construct a merged splat $m$ by matching
\emph{mass-weighted} spatial moments and applying a saturating opacity composition.
The key design choice is the weighting: rather than using opacity alone, we weight each
splat by its \emph{mass} (the integral of its normalized Gaussian density) so that
splats with larger spatial extents and/or higher opacities contribute proportionally more to
the merged moments. This is especially important under aggressive compression, where the
remaining primitives must preserve coverage without collapsing toward small, high-variance
outliers. This update has a direct geometric interpretation (Fig.~\ref{fig:pipeline}, bottom-right):
it consists of \emph{kernel averaging} (preserving typical local shape) plus \emph{kernel expansion}
(inflating uncertainty to cover between-mean dispersion). We therefore define: 

\textbf{Mass weights.}
Interpreting each splat as an unnormalized Gaussian mass density, the mass of splat $i$ is
\begin{equation}
w_i \doteq (2\pi)^{3/2}\,\alpha_i\,\prod_k s_{i,k},
\qquad
w_j \doteq (2\pi)^{3/2}\,\alpha_j\,\prod_k s_{j,k},
\qquad
W \doteq w_i + w_j .
\label{eq:vm_weights}
\end{equation}
Intuitively, $w$ approximates the total mass $\int p_i(x)\,dx$ of an unnormalized Gaussian,
and thus increases with both \emph{size} (scale) and \emph{opacity}.

\textbf{Mass-matched moments.}
The merged mean and appearance feature are mass-weighted averages:
\begin{equation}
\mu_m \;=\; \frac{w_i\mu_i + w_j\mu_j}{W}, \quad
f_m \;\doteq\; \frac{w_i f_i + w_j f_j}{W}.
\end{equation}
The merged covariance matches mass-weighted second moments:
\begin{equation}
\Sigma_m
=
\frac{
w_i\!\left(\Sigma_i + (\mu_i-\mu_m)(\mu_i-\mu_m)^\top\right)
+
w_j\!\left(\Sigma_j + (\mu_j-\mu_m)(\mu_j-\mu_m)^\top\right)
}{W}.
\label{eq:vm_cov}
\end{equation}
Equivalently, introducing $\mathcal{K}\doteq\{i,j\}$ and $W\doteq\sum_{k\in\mathcal{K}} w_k$, we obtain
\begin{equation}\label{eq:vm_cov_decomp_short}
\Sigma_m
=
\underbrace{\frac{1}{W}\sum_{k\in\mathcal{K}} w_k \,\Sigma_k}_{\text{(I) mass-weighted covariance average}}
\;+\;
\underbrace{\frac{1}{W}\sum_{k\in\mathcal{K}} w_k\,(\mu_k-\mu_m)(\mu_k-\mu_m)^\top}_{\text{(II) dispersion term}} .
\end{equation}
The first term preserves the average \emph{within-splat} shape, while the second term captures
\emph{between-mean dispersion}, expanding the merged Gaussian so that it covers the union of the two
components.

\textbf{Opacity and appearance aggregation.}
We compose opacities using the Porter--Duff \emph{source-over} (``over'') rule,
equivalently multiplying transmittances $T=1-\alpha$ 
\begin{equation}
\alpha_m \;\doteq\; 1 - (1-\alpha_i)(1-\alpha_j)
\;=\; \alpha_i + \alpha_j - \alpha_i\alpha_j .
\label{eq:vm_alpha}
\end{equation}

We refer to this operator as \textbf{Mass Preserved Moment Matching (MPMM)}. Compared to classical
moment matching and linear opacity conservation, MPMM biases
merges toward \emph{coverage} by accounting for splat volume, which empirically reduces
holes and improves rendering stability under extreme compression.

\vspace{-3mm}
\subsection{Progressive Batched Merging}
\label{sec:progmerge}

We perform simplification by repeatedly collapsing low-cost edges in the merge graph,
analogous to edge-collapse procedures in mesh simplification.

\textbf{Batched non-overlapping edge collapses.}
Given edge costs $\mathcal{C}(i,j)$ on $E$ (Eq.~\ref{Equ:cost}), each pass selects a set of
\emph{non-overlapping} low-cost edges (a matching) and merges all selected pairs in parallel
using MPMM (Sec.~\ref{sec:merge}). Collapsing edges produces a new splat set and a coarser graph.
Iterating this process yields a hierarchy of representations from fine to coarse, and we can stop
once we reach a target number of splats or target compression ratio.

\textbf{Local neighborhood refresh.}
Merges change local geometry and density, so a fixed initial KNN graph can miss newly adjacent
merge candidates. We therefore periodically refresh candidate edges by re-running KNN queries
around newly formed splats (and/or their local neighborhoods). This preserves scalability while
allowing newly relevant pairs to become eligible candidates, improving merge quality versus a
fully fixed edge set.

\vspace{-3mm}
\section{Experiments}
\vspace{-2mm}

\begin{figure}[!t]
    \centering
    \includegraphics[width=0.93\linewidth]{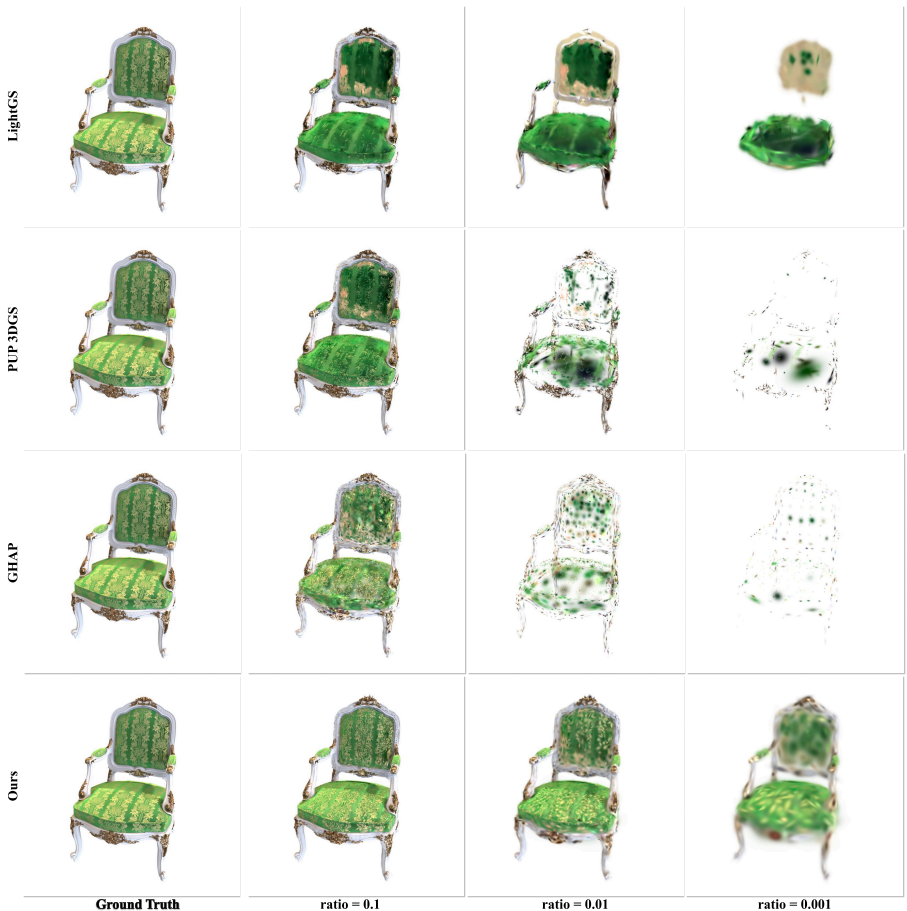}
    \caption{\textbf{Qualitative comparison on chair from NeRF Synthetic dataset.}
    We show two representative test views. Columns correspond to the compaction ratio
    $\rho\in\{0.1,0.01,0.001\}$ (leftmost: ground truth), and rows compare \cite{lightGaussian, GHAP, PUP3DGS}, and Ours.}
    \label{fig:chair}
    \vspace{-3mm}
\end{figure}

\begin{figure}
    \centering
    \includegraphics[width=\linewidth]{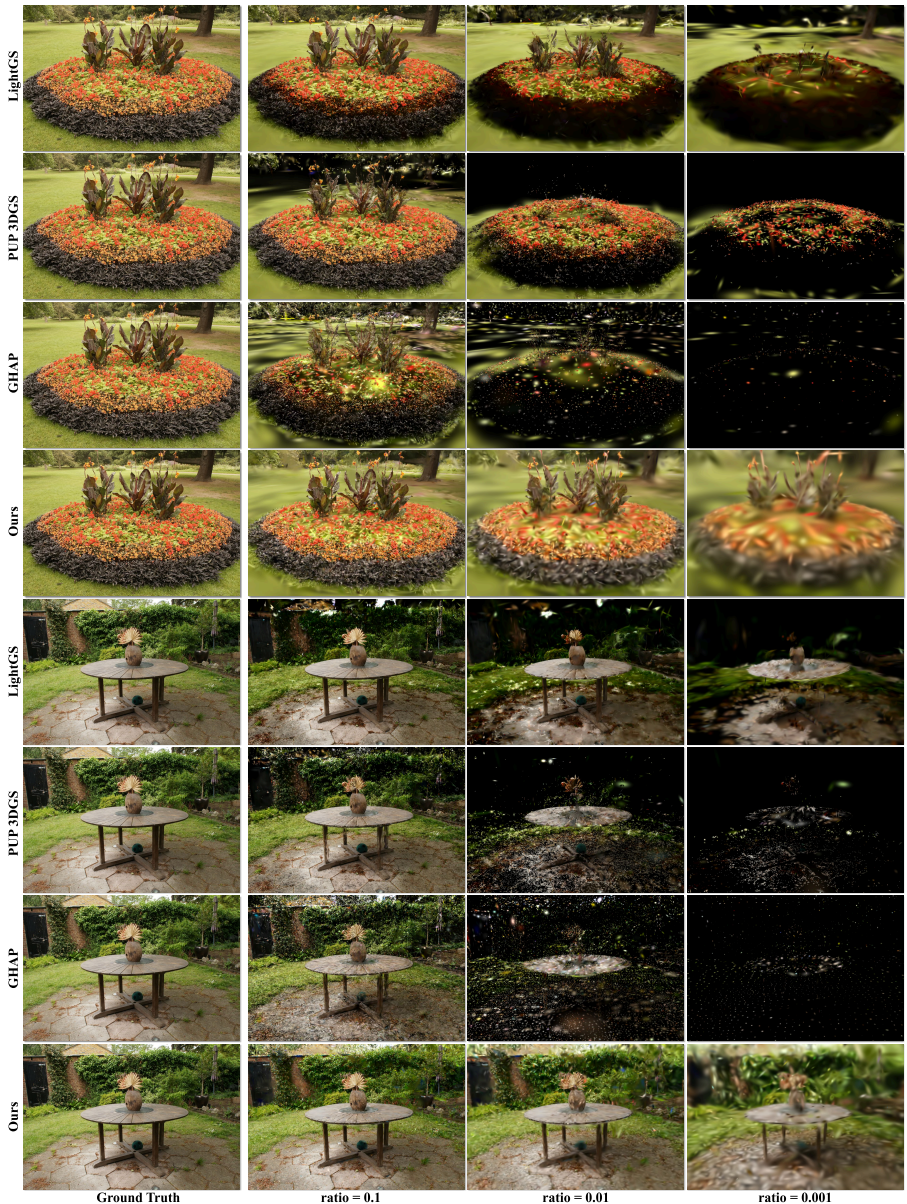}
    \caption{\textbf{Qualitative comparison on Mip-NeRF 360 dataset.}
    We show two representative test views. Columns correspond to the compaction ratio
    $\rho\in\{0.1,0.01,0.001\}$ (leftmost: ground truth), and rows compare \cite{lightGaussian, GHAP, PUP3DGS}, and Ours.}
    \label{fig:m360}
    \vspace{-4mm}
\end{figure}

\newcommand{\first}[1]{\cellcolor{red!20}#1}
\newcommand{\second}[1]{\cellcolor{pink!30}#1}
\newcommand{\third}[1]{\cellcolor{yellow!25}#1}
\begin{table*}[t]
\centering
\setlength{\tabcolsep}{4pt}
\renewcommand{\arraystretch}{1.15}
\small
\resizebox{\linewidth}{!}{%
\begin{tabular}{llccccccccc}
\toprule
\multirow{2}{*}{\shortstack{Dataset\\(PSNR/SSIM/FPS)\\Full}} & \multirow{2}{*}{Method} & \multicolumn{3}{c}{$\rho = 0.1$} & \multicolumn{3}{c}{$\rho = 0.01$} & \multicolumn{3}{c}{$\rho = 0.001$} \\
\cmidrule(lr){3-5} \cmidrule(lr){6-8} \cmidrule(lr){9-11}
& & PSNR & SSIM & FPS & PSNR & SSIM & FPS & PSNR & SSIM & FPS \\
\midrule
\multirow{4}{*}{\shortstack{NeRF Synth\cite{mildenhall2021nerf}\\33.66/0.970/854.04}} 
& LightGS~\cite{lightGaussian} & \second{21.25} & \second{0.875} & 2204.97 & \second{15.76} & \second{0.807} & 2571.07 & \second{12.64} & \second{0.796} & 2623.73 \\
& PUP3DGS~\cite{PUP3DGS} & \third{20.24} & \third{0.860} & 2341.34 & \third{13.26} & \third{0.786} & 2676.07 & \third{11.31} & \third{0.792} & 2838.60 \\
& GHAP~\cite{GHAP} & 20.09 & 0.838 & 2486.94 & 12.81 & 0.776 & 2679.19 & 11.07 & 0.791 & 2916.30 \\
& Ours & \first{25.81} & \first{0.910} & 2450.17 & \first{22.28} & \first{0.858} & 2598.03 & \first{19.04} & \first{0.822} & 2641.46 \\
\midrule
\multirow{4}{*}{\shortstack{Mips-360\cite{barron2021mipnerf}\\27.46/0.815/182.43}}
& LightGS~\cite{lightGaussian} & \second{19.38} & \first{0.588} & 727.19 & \second{14.39} & \second{0.389} & 1525.92 & \second{11.97} & \second{0.278} & 2070.79 \\
& PUP3DGS~\cite{PUP3DGS} & 15.59 & \third{0.513} & 996.41 & 10.19 & 0.161 & 2407.68 & \third{8.81} & \third{0.056} & 2650.41 \\
& GHAP~\cite{GHAP} & \third{17.35} & 0.444 & 1224.46 & \third{10.62} & \third{0.174} & 2557.02 & 8.52 & 0.035 & 2622.83 \\
& Ours & \first{21.97} & \second{0.582} & 1015.20 & \first{19.39} & \first{0.470} & 2005.76 & \first{17.20} & \first{0.430} & 2193.87 \\
\midrule
\multirow{4}{*}{\shortstack{T\&T\cite{tanksandtemples}\\23.63/0.847/299.07}}
& LightGS~\cite{lightGaussian} & \second{17.50} & \first{0.642} & 1108.22 & \second{12.29} & \second{0.441} & 1995.98 & \second{9.36} & \second{0.341} & 2418.99 \\
& PUP3DGS~\cite{PUP3DGS} & 12.71 & \third{0.564} & 1283.38 & 8.22 & \third{0.266} & 2433.64 & \third{6.78} & \third{0.162} & 2593.01 \\
& GHAP~\cite{GHAP} & \third{15.34} & 0.487 & 1665.64 & \third{8.43} & 0.220 & 2659.11 & 5.33 & 0.026 & 2670.40 \\
& Ours & \first{17.94} & \second{0.626} & 1429.88 & \first{15.29} & \first{0.501} & 2325.87 & \first{13.54} & \first{0.457} & 2487.31 \\
\midrule
\multirow{4}{*}{\shortstack{D\&B\cite{deepblending}\\29.56/0.903/233.25}}
& LightGS~\cite{lightGaussian} & \second{24.28} & \second{0.816} & 891.18 & \second{18.28} & \second{0.712} & 1878.64 & \second{13.41} & \second{0.616} & 2295.95 \\
& PUP3DGS~\cite{PUP3DGS} & 19.65 & \third{0.755} & 1325.03 & 8.83 & 0.282 & 2478.12 & \third{7.06} & \third{0.073} & 2672.21 \\
& GHAP~\cite{GHAP} & \third{21.75} & 0.739 & 1375.59 & \third{11.36} & \third{0.435} & 2583.11 & \third{7.06} & 0.061 & 2627.05 \\
& Ours & \first{26.29} & \first{0.839} & 1202.23 & \first{23.12} & \first{0.780} & 2132.31 & \first{19.42} & \first{0.739} & 2321.92 \\
\bottomrule
\end{tabular}%
}
\caption{Quantitative benchmarks across datasets and sampling ratios $\rho$. Full-quality references (PSNR/SSIM/FPS at $\rho=1.0$) are listed in the Dataset column. For $\rho\in\{0.1,0.01,0.001\}$, color indicates rank within each dataset/metric column: \colorbox{red!20}{red}=1st, \colorbox{pink!30}{pink}=2nd, \colorbox{yellow!25}{yellow}=3rd.}
\label{tab:Benchmarks}
\vspace{-6mm}
\end{table*}

We evaluate on four widely used 3DGS benchmarks spanning both synthetic and real captured scenes: NeRF-Synthetic~\cite{mildenhall2021nerf} (8 scenes), Mip-NeRF 360~\cite{mip360} (9 unbounded real-world scenes), Tanks and Temples~\cite{tanksandtemples} (2 large-scale real scenes), and Deep Blending~\cite{deepblending} (2 challenging indoor scenes). This combination covers diverse geometry, appearance complexity, and camera trajectories.

\textbf{Evaluation Metrics.}
We report Peak Signal-to-Noise Ratio
(PSNR) and Structural Similarity Index Measure (SSIM) as primary visual fidelity metrics, and also report rendering speed (FPS) in Table~\ref{tab:Benchmarks}. All metrics are computed on held-out test views following each benchmark's standard evaluation protocol.

\textbf{Comparison Protocol.}
Table~\ref{tab:Benchmarks} compares against LightGS~\cite{lightGaussian}, PUP3DGS~\cite{PUP3DGS}, and GHAP~\cite{GHAP} at three representative budgets, $\rho\in\{0.1,0.01,0.001\}$: $\rho=0.1$ follows the common GHAP setting, while $\rho=0.01$ and $\rho=0.001$ reflect more aggressive LOD-style compression. Since our objective is \emph{training-free} post-hoc compaction, we do not perform any post-compression fine-tuning; for methods typically paired with a fine-tuning stage (e.g., LightGS and PUP3DGS), we evaluate only their pruning/selection stage to isolate compaction quality.

\textbf{Implementation Details.}
Our method operates directly on pretrained 3DGS assets and preserves the original 3DGS representation, so outputs remain compatible with standard Gaussian splatting renderers. Monte Carlo merge-cost estimation uses $S{=}1$ sample by default ($S{=}128$ yields no noticeable quality gain); we set the KNN graph size to $K{=}16$, merge $N/2$ disjoint pairs per pass, and refresh the graph each pass. Unless otherwise noted, all experiments use a single NVIDIA RTX 4090 GPU with CUDA 12.8, with FPS measured on the same machine, and we report final results at each target ratio without retraining.

\vspace{-4mm}
\subsection{Comparison with the State of the Art}
\vspace{-2mm}

Qualitatively (Fig.~\ref{fig:m360}, Fig.~\ref{fig:chair}), all methods are broadly faithful at moderate compression ($\rho{=}0.1$). As the budget decreases ($\rho{=}0.01, 0.001$), baselines collapse into sparse high-intensity splats and lose coherent surfaces, consistent with unstable merges and floater-dominated candidates, whereas our filtering and mass-preserving moment matching maintain contiguous coverage and degrade gracefully. Additional comparisons are in the appendix.

Quantitatively (Table~\ref{tab:Benchmarks}), our method achieves the best PSNR at every budget across all four benchmarks, and its advantage widens as compression becomes more aggressive: averaged over datasets, PSNR improves over the best baseline by \textbf{+2.40 dB} at $\rho{=}0.1$, \textbf{+4.84 dB} at $\rho{=}0.01$, and \textbf{+5.46 dB} at $\rho{=}0.001$. Gains are largest on NeRF-Synthetic (up to \textbf{+6.52 dB}) and grow consistently with compression on the real-scene benchmarks (Mip-NeRF 360, Tanks \& Temples, Deep Blending), where they reflect fewer missing-geometry and floater failures. Overall, baselines degrade rapidly as $\rho$ decreases while our merge-graph pipeline retains substantially higher fidelity, most visibly in the most aggressive regimes.
\vspace{-4mm}
\subsection{Runtime and Storage}
\vspace{-2mm}
\label{sec:runtime_storage}

\begin{table}[t]
\centering
\begin{minipage}[t]{0.40\columnwidth}
\centering
\caption{Wall-clock runtime to compact to $\rho{=}0.1$, CPU vs.\ single GPU.}
\label{Tab:wall_time}
\resizebox{\columnwidth}{!}{
\begin{tabular}{lccc}
\toprule
\textbf{Dataset} & \textbf{CPU} & \textbf{GPU} & \textbf{Orig.} \\
 & \textbf{(s)} & \textbf{(s)} & \textbf{(MB)} \\
\midrule
M360 Out  & 88.10 & 17.14 & 1122 \\
M360 In   & 25.27 & 5.07  & 353 \\
T \& T    & 19.41 & 4.12  & 255 \\
D \& B    & 51.69 & 9.75  & 663 \\
NeRF-Syn  & 4.57  & 1.63  & 67 \\
\bottomrule
\end{tabular}
}
\end{minipage}
\hfill
\begin{minipage}[t]{0.58\columnwidth}
\centering
\caption{Combined storage (MB): NanoGS $\rightarrow$ SOG~\cite{SOG}. Cells: compacted $\rightarrow$ compressed ($\times$ from SOG).}
\label{Tab:disk_size}
\resizebox{\columnwidth}{!}{
\begin{tabular}{lccc}
\toprule
\textbf{Dataset} & \textbf{$r{=}0.1$} & \textbf{$r{=}0.01$} & \textbf{$r{=}0.001$} \\
\midrule
M360 Out   & 112.23$\rightarrow$16.07 & 11.22$\rightarrow$1.61 & 1.12$\rightarrow$0.18 \\
M360 In    & 34.29$\rightarrow$4.80   & 3.43$\rightarrow$0.48  & 0.34$\rightarrow$0.06 \\
T \& T     & 25.48$\rightarrow$3.68   & 2.55$\rightarrow$0.39  & 0.26$\rightarrow$0.04 \\
D \& B     & 66.27$\rightarrow$9.95   & 6.63$\rightarrow$1.02  & 0.66$\rightarrow$0.11 \\
NeRF-Syn   & 6.69$\rightarrow$1.11    & 0.67$\rightarrow$0.11  & 0.07$\rightarrow$0.01 \\
\midrule
\textbf{Mean} & \multicolumn{3}{c}{16.83$\rightarrow$2.45 (6.35$\times$), 60 models} \\
\bottomrule
\end{tabular}
}
\end{minipage}
\vspace{-4mm}
\end{table}

NanoGS is lightweight enough to run on CPU. Table~\ref{Tab:wall_time} reports wall-clock time to compact to $\rho{=}0.1$: even the largest outdoor Mip-NeRF~360 scenes finish in under $90$\,s on CPU and under $20$\,s on a single GPU, with smaller scenes completing in seconds. Because NanoGS preserves the standard 3DGS representation, it composes with orthogonal bit-level compression: Table~\ref{Tab:disk_size} chains our compaction with Self-Organized Gaussians (SOG)~\cite{SOG}, yielding a further $\sim\!6.3\times$ mean storage reduction on top of compaction across all $60$ models, for a combined reduction of several orders of magnitude at the most aggressive ratios.
\vspace{-4mm}
\subsection{Ablation Study}
\vspace{-2mm}
\noindent

Table~\ref{tab:ablation} evaluates four variants:
\textbf{w/o KNN graph} replaces KNN-graph candidate selection with Octree-based neighborhood selection,
\textbf{w/o filtering} disables the initial opacity-based pruning,
\textbf{w/o I-divergence} replaces the I-divergence geometric term with a fast Mean Squared Error (MSE) surrogate,
and \textbf{Full} uses the complete formulation.
\begin{table}[htbp]
\centering
\setlength{\tabcolsep}{4pt}
\renewcommand{\arraystretch}{1.15}
\small
\begin{tabular}{llcccccc}
\toprule
\multirow{2}{*}{Dataset} & \multirow{2}{*}{Method} & \multicolumn{2}{c}{$\rho = 0.1$} & \multicolumn{2}{c}{$\rho = 0.01$} & \multicolumn{2}{c}{$\rho = 0.001$} \\
\cmidrule(lr){3-4} \cmidrule(lr){5-6} \cmidrule(lr){7-8}
& & PSNR & SSIM & PSNR & SSIM & PSNR & SSIM \\
\midrule
\multirow{4}{*}{NeRF}
 & w/o KNN graph & 25.53 & 0.905 & 20.74 & 0.839 & 15.85 & 0.800 \\
 & w/o filtering & 24.99 & 0.904 & 22.01 & \textbf{0.858} & 18.87 & \textbf{0.822} \\
 & w/o I-divergence & 25.65 & 0.908 & 22.15 & 0.854 & 18.80 & 0.819 \\
 & Full & \textbf{25.81} & \textbf{0.910} & \textbf{22.28} & \textbf{0.858} & \textbf{19.04} & \textbf{0.822} \\
\midrule
\multirow{4}{*}{M360}
 & w/o KNN graph & 21.53 & 0.537 & 17.82 & 0.444 & 14.11 & 0.382 \\
 & w/o filtering & 18.24 & 0.504 & 15.47 & 0.421 & 14.67 & 0.401 \\
 & w/o I-divergence & 21.61 & 0.566 & 19.19 & 0.467 & 16.98 & 0.426 \\
 & Full & \textbf{21.97} & \textbf{0.582} & \textbf{19.39} & \textbf{0.470} & \textbf{17.20} & \textbf{0.430} \\
\bottomrule
\end{tabular}
\caption{Ablation study across datasets and compaction ratios. \textbf{w/o KNN graph} replaces the KNN-graph candidate selection with the Octree-based one; \textbf{w/o filtering} disables opacity-based filtering; \textbf{w/o I-divergence} replaces the principled I-divergence geometric cost with an Mean Squared Error surrogate; \textbf{Full} uses the complete proposed formulation. Bold indicates the best result within each dataset/metric column.}
\label{tab:ablation}
\vspace{-4mm}
\end{table}

\textbf{Candidate selection.}
Replacing the KNN-graph candidate construction with Octree-based selection leads to a clear quality drop, especially at aggressive budgets: on Mip-NeRF 360 at $\rho{=}0.001$, PSNR drops from \textbf{17.20} (Full) to \textbf{14.11}, and on NeRF-Synthetic from \textbf{19.04} to \textbf{15.85}, indicating that KNN-graph locality provides more reliable merge candidates than the Octree alternative.

\textbf{Opacity filtering.}
Removing filtering consistently hurts performance, particularly on Mip-NeRF 360 (\textbf{17.20}$\rightarrow$\textbf{14.67} at $\rho{=}0.001$), while NeRF-Synthetic is less sensitive, suggesting that filtering suppresses the low-opacity ``floaters'' common in real-world reconstructions.

\textbf{Merge cost.}
Replacing the I-divergence geometric distortion with the fast surrogate consistently degrades quality across both datasets and all ratios, indicating that the principled I-divergence formulation is a more faithful geometric fidelity measure than the MSE approximation.

In sum, KNN-graph candidate selection improves merge reliability, filtering removes noisy floaters early, and the I-divergence cost adds geometry-aware guidance. Notably, the first and third of these are stage swaps---topology and cost objective---whose effect on quality directly demonstrates that the stages are independently consequential.

\vspace{-2mm}
\section{Conclusion and Discussion}
\label{sec:discussion}
\vspace{-2mm}
We presented \textbf{NanoGS}, a training-free, CPU-friendly framework for Gaussian Splat simplification that operates directly on pretrained 3DGS models without calibrated images or scene-specific optimization. By formulating simplification as progressive local pairwise merging over a sparse $k$-nearest-neighbor graph, NanoGS avoids the GPU-intensive retraining pipelines that limit existing compaction methods, and our GPU implementation of the graph-construction and edge-collapse stages further reduces wall-clock time by roughly $5\times$ (Table~\ref{Tab:wall_time}). Across four standard benchmarks and varying compression budgets, it consistently outperforms state-of-the-art compaction methods, with particularly stable degradation at extreme ratios where prior methods collapse into floater-dominated artifacts. Crucially, NanoGS is not merely a merging algorithm but a \emph{structural system}: rather than forcing merges through a rigid spatial grid, it decouples candidate selection, merge cost, and the merge operator into independent, interchangeable stages, so that strong compaction follows from the structure rather than any single component---and we regard this framework, which our ablations exercise along the topology and cost axes, as the central contribution. Each stage is also a substitution point that opens future directions: the analytic cost could be replaced by a learned neural criterion capturing perceptually or semantically important splats; the operator could extend beyond pairwise fusion to many-to-one merging, or, by \emph{inverting} from compression to generation, become a super-resolution procedure that splits primitives to add detail. Finally, extending the framework from static to dynamic 3DGS representations is a natural and practically important next step.

\section{Acknowledgment}
The project or effort depicted was sponsored by the U.S. Army Combat Capabilities Development Command under contract number W912CG-24-D-0001 and a research agreement with Coolant Climate, Inc. The content of the information does not necessarily reflect the position or the policy of the Government, and no official endorsement should be inferred.


%
%
\bibliographystyle{splncs04}
\bibliography{main}

\maketitlesupplementary
\appendix
\section{Special Case for I Divergence}
\label{app:I-Divergence}

\paragraph{Analytical cases and practical estimator.}
For two \emph{single} unnormalized Gaussians
$p_i(x)=m_i\,\mathcal{N}(x\mid\mu_i,\Sigma_i)$ and
$p_j(x)=m_j\,\mathcal{N}(x\mid\mu_j,\Sigma_j)$,
the I-divergence
\begin{equation}
D(p_i\|p_j)
=
\int \left[p_i(x)\log\frac{p_i(x)}{p_j(x)}-p_i(x)+p_j(x)\right]dx
\end{equation}
admits a closed-form expression
\begin{equation}
D(p_i\|p_j)
=
m_i\Big(\log\frac{m_i}{m_j}+\mathrm{KL}(\mathcal{N}_i\|\mathcal{N}_j)\Big)
-m_i+m_j,
\end{equation}
where
\begin{equation}
\mathrm{KL}(\mathcal{N}_i\|\mathcal{N}_j)
=
\frac{1}{2}\Big(
\mathrm{tr}(\Sigma_j^{-1}\Sigma_i)
+
(\mu_j-\mu_i)^{\top}\Sigma_j^{-1}(\mu_j-\mu_i)
-d
+
\log\frac{\det\Sigma_j}{\det\Sigma_i}
\Big),\qquad d=3.
\end{equation}
This analytical form is commonly used in Gaussian mixture reduction when both
distributions belong to the Gaussian family.

However, the distortion used in our merge criterion compares a
\emph{two-component mixture} with a \emph{single Gaussian}:
\begin{equation}
\mathcal{D}_{\mathrm{geo}}(i,j)
=
\mathrm{KL}\!\left(\tilde p_{ij}\,\middle\|\, q_m\right),
\qquad
\tilde p_{ij}(x)=\pi\,\mathcal{N}_i(x)+(1-\pi)\,\mathcal{N}_j(x),
\end{equation}
where $q_m=\mathcal{N}(x;\mu_m,\Sigma_m)$ is the merged Gaussian obtained
by moment matching.
Because $\tilde p_{ij}$ is a Gaussian \emph{mixture}, the term
$\log(\pi\mathcal{N}_i(x)+(1-\pi)\mathcal{N}_j(x))$
prevents the integral from admitting a closed-form expression.
Consequently, the analytical formulas above cannot be directly applied.

To evaluate the distortion efficiently, we instead approximate the KL divergence
using Monte Carlo sampling:
\begin{equation}
\mathrm{KL}\!\left(\tilde p_{ij}\,\middle\|\, q_m\right)
=
\mathbb{E}_{x\sim \tilde p_{ij}}\!\left[
\log\!\big(\pi\,\mathcal{N}_i(x)+(1-\pi)\,\mathcal{N}_j(x)\big)
-
\log q_m(x)
\right].
\end{equation}
Given samples $\{x_s\}_{s=1}^S\sim \tilde p_{ij}$, we estimate
\begin{equation}
\mathrm{KL}\!\left(\tilde p_{ij}\,\middle\|\, q_m\right)
\approx
\frac{1}{S}\sum_{s=1}^S
\left[
\log\!\big(\pi\,\mathcal{N}_i(x_s)+(1-\pi)\,\mathcal{N}_j(x_s)\big)
-
\log q_m(x_s)
\right].
\label{eq:mc_kl_estimator}
\end{equation}

Sampling from $\tilde p_{ij}$ is straightforward: we draw from component $i$
with probability $\pi$ and from component $j$ with probability $1-\pi$.
In practice we use a small fixed number of samples $S$ and share common random
numbers across candidate pairs to reduce variance while keeping the computation
lightweight. This Monte Carlo estimator is the procedure implemented in our
codebase.

\section{Algorithms}
\label{App:Algorithm}

Algorithm~\ref{alg:knn_full_merge_impl} describes the greedy KNN edge-collapse procedure used in our implementation. The algorithm takes as input an initial set of splats and a target keep ratio, and progressively reduces the representation through repeated local merges.

The procedure begins with an opacity filtering stage. Since extremely low-opacity splats contribute little to the rendered scene, removing them early improves efficiency and reduces the search space for subsequent merge operations. After this preprocessing step, the algorithm enters an iterative simplification loop.

At each iteration, we build a sparse $k$-nearest-neighbor graph over the current splat centers. This graph serves as a locality prior: instead of considering all pairwise combinations, we only evaluate merges between nearby splats. The resulting candidate set is converted into a unique undirected edge list, and each edge is scored by the proposed merge cost. This cost measures how much information would be lost if the two endpoint splats were replaced by a single merged splat.

The core decision step is a greedy disjoint-pair selection. Candidate edges are sorted by increasing cost, and the algorithm scans them in order, selecting an edge whenever neither endpoint has already been assigned in the current pass. This produces a set of non-overlapping merges that can be applied in parallel. The greedy rule is simple but effective: it prioritizes the cheapest local collapses first, while ensuring that the resulting update is consistent.

After the selected pairs are merged, the splat attributes are updated and the algorithm repeats on the new reduced set. Because the KNN graph and edge costs are recomputed after each pass, the method adapts to the evolving geometry of the simplified representation. The loop terminates once the target number of splats is reached or when no further valid merge is available.

Overall, this algorithm can be viewed as a practical graph coarsening strategy for Gaussian splats: the KNN graph defines feasible local collapses, the merge cost ranks them by distortion, and the greedy disjoint matching step realizes a scalable approximation to global simplification.

\begin{algorithm}[t]
\caption{Greedy KNN edge-collapse splat simplification (implementation).}
\label{alg:knn_full_merge_impl}
\KwIn{Input PLY splats $(\mu,\alpha,s,q,f)$ with $N_0$ entries; keep ratio $r\in(0,1)$; KNN size $k$; opacity prune threshold $\tau$; cost params $\mathrm{cp}$; edge-cost block size $B$.}
\KwOut{Simplified PLY with $\approx \lceil rN_0\rceil$ splats.}

Read PLY $\rightarrow (\mu,\alpha,s,q,f)$\;
$N_0 \gets |\mu|$, $N_{\text{tgt}} \gets \max(\lceil rN_0\rceil,1)$\;
$\tau' \gets \min(\tau,\mathrm{median}(\alpha))$\tcp*{robust prune threshold}
$\mathcal{I}\gets\{i:\alpha_i\ge\tau'\}$; subset $(\mu,\alpha,s,q,f)\leftarrow(\mu,\alpha,s,q,f)[\mathcal{I}]$\;

\While{$|\mu| > N_{\text{tgt}}$}{
    $N \gets |\mu|$; $k_{\text{eff}} \gets \min(\max(1,k),N-1)$\;
    $\mathrm{nbr} \gets \textsc{KNN}(\mu,k_{\text{eff}})$\tcp*{directed $k$NN per node}
    $E \gets \textsc{UndirectedUnionEdges}(\mathrm{nbr})$\tcp*{unique edges $\{i,j\}$ with $i<j$}
    $w \gets \textsc{EdgeCosts}(E,\mu,s,q,\alpha,f,\mathrm{cp},B)$\tcp*{blockwise; uses \texttt{fast\_cost\_pairs}}
    $P \gets N - N_{\text{tgt}}$\tcp*{merges needed this pass}
    $\mathcal{M} \gets \textsc{GreedyDisjointPairs}(E,w,N,P)$\tcp*{sort by $w$, pick disjoint edges}
    \If{$|\mathcal{M}|=0$}{
        \textbf{break}\;
    }
    $(\mu,s,q,\alpha,f) \gets \textsc{MergePairs}(\mu,s,q,\alpha,f,\mathcal{M})$\;
}

$\alpha \gets \mathrm{clip}(\alpha,0,1)$; write output PLY\;
\end{algorithm}

\section{Additional Experiment Results}
We provide per-scene quantitative results for all four datasets in
Tab.~\ref{Tab:per_scene_deep_blending},
Tab.~\ref{Tab:per_scene_mips360},
Tab.~\ref{Tab:per_scene_nerf_synthetic}, and
Tab.~\ref{Tab:per_scene_tanj_templates}.
\begin{table*}[t]
\centering
\setlength{\tabcolsep}{4pt}
\renewcommand{\arraystretch}{1.15}
\small
\resizebox{\linewidth}{!}{%
\begin{tabular}{lccccccccc}
\toprule
\multirow{2}{*}{Scene} & \multicolumn{3}{c}{$\rho = 0.1$} & \multicolumn{3}{c}{$\rho = 0.01$} & \multicolumn{3}{c}{$\rho = 0.001$} \\
\cmidrule(lr){2-4} \cmidrule(lr){5-7} \cmidrule(lr){8-10}
& PSNR $\uparrow$ & SSIM $\uparrow$ & LPIPS $\downarrow$ & PSNR $\uparrow$ & SSIM $\uparrow$ & LPIPS $\downarrow$ & PSNR $\uparrow$ & SSIM $\uparrow$ & LPIPS $\downarrow$ \\
\midrule
\rowcolor{gray!10} \multicolumn{10}{l}{\textit{Outdoor Scenes}} \\
bicycle & 20.1650 & 0.4570 & 0.4558 & 18.6070 & 0.3469 & 0.6292 & 17.5477 & 0.3152 & 0.7256 \\
flowers & 17.6026 & 0.3619 & 0.5214 & 16.0689 & 0.2604 & 0.6744 & 14.9127 & 0.2324 & 0.7607 \\
garden  & 21.0113 & 0.4934 & 0.4052 & 19.0978 & 0.3418 & 0.5917 & 16.7483 & 0.3056 & 0.7366 \\
stump   & 21.0693 & 0.4614 & 0.4626 & 19.5957 & 0.3685 & 0.6364 & 18.8344 & 0.3500 & 0.7095 \\
treehill& 19.0771 & 0.4168 & 0.5460 & 18.0260 & 0.3766 & 0.6566 & 17.0083 & 0.3608 & 0.6870 \\
\midrule
\rowcolor{gray!10} \multicolumn{10}{l}{\textit{Indoor Scenes}} \\
room    & 26.1853 & 0.8025 & 0.3730 & 22.3240 & 0.7219 & 0.4879 & 18.3340 & 0.6526 & 0.5312 \\
counter & 23.5919 & 0.7390 & 0.3838 & 20.2366 & 0.6447 & 0.5217 & 17.0012 & 0.5884 & 0.5723 \\
kitchen & 24.0360 & 0.7187 & 0.3549 & 19.9008 & 0.5184 & 0.5631 & 17.1972 & 0.4791 & 0.6530 \\
bonsai  & 24.9709 & 0.7831 & 0.3849 & 20.6622 & 0.6484 & 0.5192 & 17.1862 & 0.5830 & 0.5716 \\
\bottomrule
\end{tabular}%
}
\caption{Quantitative results for Mip-NeRF 360 dataset across three compaction ratios.}
\label{Tab:per_scene_mips360}
\end{table*}

\begin{table*}[t]
\centering
\setlength{\tabcolsep}{4pt}
\renewcommand{\arraystretch}{1.15}
\small
\resizebox{\linewidth}{!}{%
\begin{tabular}{lccccccccc}
\toprule
\multirow{2}{*}{Scene} & \multicolumn{3}{c}{$\rho = 0.1$} & \multicolumn{3}{c}{$\rho = 0.01$} & \multicolumn{3}{c}{$\rho = 0.001$} \\
\cmidrule(lr){2-4} \cmidrule(lr){5-7} \cmidrule(lr){8-10}
& PSNR $\uparrow$ & SSIM $\uparrow$ & LPIPS $\downarrow$ & PSNR $\uparrow$ & SSIM $\uparrow$ & LPIPS $\downarrow$ & PSNR $\uparrow$ & SSIM $\uparrow$ & LPIPS $\downarrow$ \\
\midrule
drums     & 23.5546 & 0.9139 & 0.0783 & 20.6873 & 0.8580 & 0.1427 & 17.4901 & 0.8112 & 0.2129 \\
lego      & 22.5217 & 0.8791 & 0.1173 & 20.4297 & 0.8036 & 0.1947 & 18.1469 & 0.7763 & 0.2500 \\
materials & 24.2313 & 0.8992 & 0.1007 & 20.5123 & 0.8360 & 0.1649 & 16.6823 & 0.7926 & 0.2343 \\
mic       & 27.5273 & 0.9632 & 0.0390 & 23.8959 & 0.9291 & 0.0789 & 20.4707 & 0.9014 & 0.1145 \\
ship      & 22.3706 & 0.7958 & 0.2129 & 19.4858 & 0.7432 & 0.3000 & 16.3459 & 0.7061 & 0.3567 \\
chair     & 27.2220 & 0.9256 & 0.0624 & 24.0736 & 0.8772 & 0.1193 & 21.7324 & 0.8554 & 0.1625 \\
ficus     & 29.4857 & 0.9634 & 0.0336 & 23.6854 & 0.9105 & 0.0804 & 20.6924 & 0.8656 & 0.1373 \\
hotdog    & 29.5549 & 0.9398 & 0.0900 & 25.4385 & 0.9037 & 0.1444 & 20.7310 & 0.8695 & 0.1893 \\
\bottomrule
\end{tabular}%
}
\caption{Quantitative results for NeRF Synthetic dataset across three compaction ratios.}
\label{Tab:per_scene_nerf_synthetic}
\end{table*}

\begin{table*}[t]
\centering
\setlength{\tabcolsep}{4pt}
\renewcommand{\arraystretch}{1.15}
\small
\resizebox{\linewidth}{!}{%
\begin{tabular}{lccccccccc}
\toprule
\multirow{2}{*}{Scene} & \multicolumn{3}{c}{$\rho = 0.1$} & \multicolumn{3}{c}{$\rho = 0.01$} & \multicolumn{3}{c}{$\rho = 0.001$} \\
\cmidrule(lr){2-4} \cmidrule(lr){5-7} \cmidrule(lr){8-10}
& PSNR $\uparrow$ & SSIM $\uparrow$ & LPIPS $\downarrow$ & PSNR $\uparrow$ & SSIM $\uparrow$ & LPIPS $\downarrow$ & PSNR $\uparrow$ & SSIM $\uparrow$ & LPIPS $\downarrow$ \\
\midrule
truck & 19.4503 & 0.6728 & 0.3851 & 16.7092 & 0.5401 & 0.5539 & 14.3262 & 0.4821 & 0.6515 \\
train & 16.4383 & 0.5800 & 0.4416 & 13.8683 & 0.4617 & 0.5975 & 12.7557 & 0.4311 & 0.6304 \\
\bottomrule
\end{tabular}%
}
\caption{Quantitative results for Tanks and Temples dataset across three compaction ratios.}
\label{Tab:per_scene_tanj_templates}
\end{table*}

\begin{table*}[t]
\centering
\setlength{\tabcolsep}{4pt}
\renewcommand{\arraystretch}{1.15}
\small
\resizebox{\linewidth}{!}{%
\begin{tabular}{lccccccccc}
\toprule
\multirow{2}{*}{Scene} & \multicolumn{3}{c}{$\rho = 0.1$} & \multicolumn{3}{c}{$\rho = 0.01$} & \multicolumn{3}{c}{$\rho = 0.001$} \\
\cmidrule(lr){2-4} \cmidrule(lr){5-7} \cmidrule(lr){8-10}
& PSNR $\uparrow$ & SSIM $\uparrow$ & LPIPS $\downarrow$ & PSNR $\uparrow$ & SSIM $\uparrow$ & LPIPS $\downarrow$ & PSNR $\uparrow$ & SSIM $\uparrow$ & LPIPS $\downarrow$ \\
\midrule
drjohnson & 25.5296 & 0.8241 & 0.3879 & 22.7212 & 0.7618 & 0.4872 & 19.6485 & 0.7265 & 0.5159 \\
playroom  & 27.0528 & 0.8536 & 0.3545 & 23.5220 & 0.7989 & 0.4473 & 19.2004 & 0.7519 & 0.4986 \\
\bottomrule
\end{tabular}%
}
\caption{Quantitative results for Deep Blending dataset across three compaction ratios.}
\label{Tab:per_scene_deep_blending}
\end{table*}

In addition to these quantitative results, we also include a demo qualitative
video in the supplementary material, which provides further visual comparisons
across scenes and compression settings. In the demo video, we also display for large scale out door scene, usually takes about over 10Gb memory, how our compression algorithms perform. 

\section{Generation Results}
\begin{figure}
    \centering
    \includegraphics[width=0.9\linewidth]{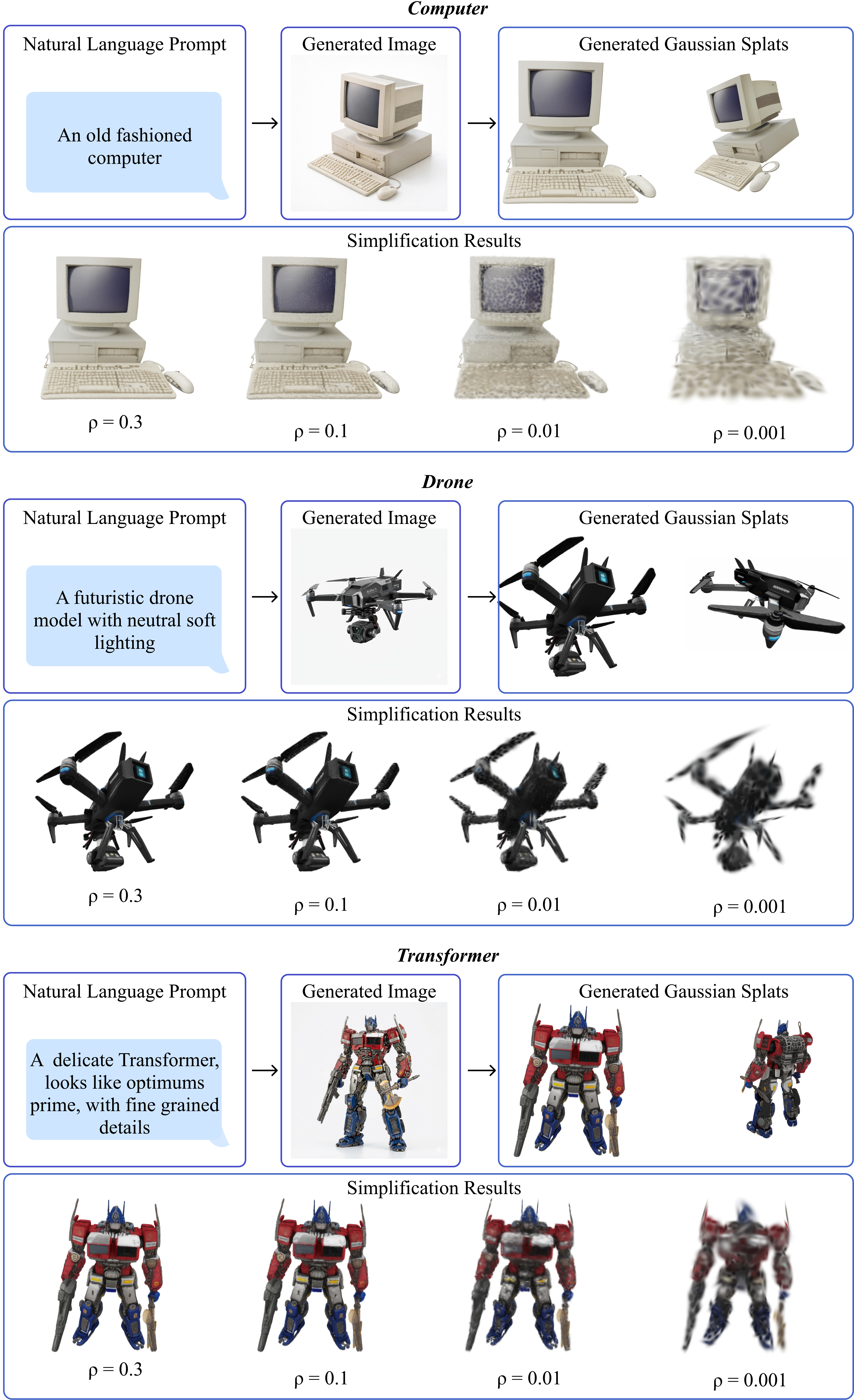}
    \caption{Generated results and corresponding compression results}
    \label{fig:Generation}
\end{figure}

\begin{figure}
    \centering
    \includegraphics[width=0.9\linewidth]{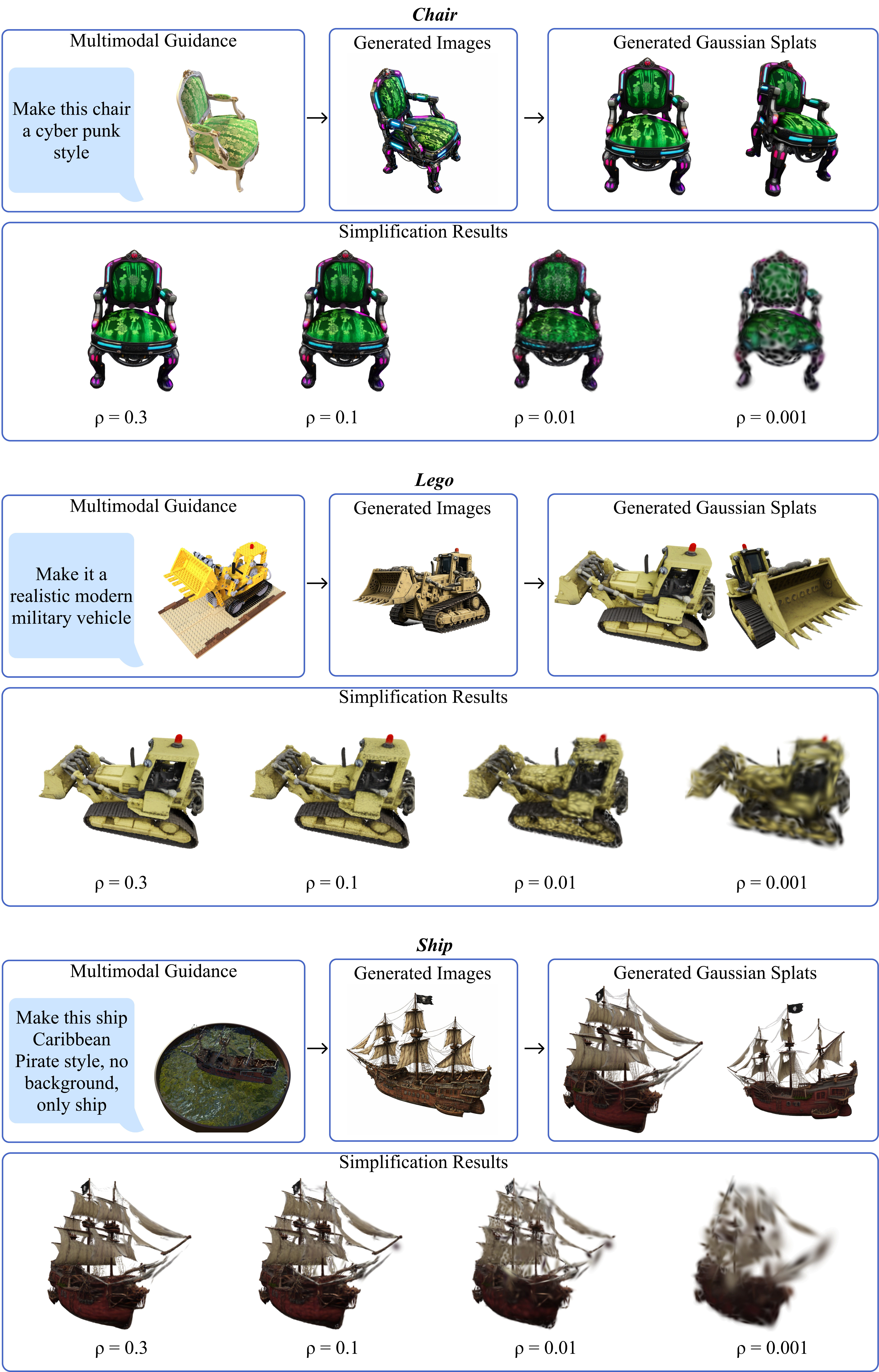}
    \caption{Generated results and corresponding compression results with multimodal inputs}
    \label{fig:Generation_image}
\end{figure}

As stated in the introduction and related work, current 3D Gaussian generation pipelines do not guarantee consistent camera poses or image fidelity. To verify this, we conducted additional experiments on generated data using the following pipeline:

First, we use natural-language prompts to guide mainstream image generation models, such as ChatGPT\footnote{OpenAI, ``ChatGPT.'' \url{https://chatgpt.com/} (accessed March 12, 2026).} and Gemini\footnote{Google, ``Gemini.'' \url{https://gemini.google.com/} (accessed March 12, 2026).}. The resulting images are then used as input for Trellis\footnote{Jianfeng Xiang \emph{et al.}, ``Structured 3D Latents for Scalable and Versatile 3D Generation,'' in \emph{Proceedings of the IEEE/CVF Conference on Computer Vision and Pattern Recognition (CVPR)}, 2025.} to generate 3D Gaussian Splats. Finally, we apply our compression method to these assets. The results are illustrated in Fig.~\ref{fig:Generation}. The prompts used for 2D image generation can also be multimodal, as shown in Fig.~\ref{fig:Generation_image}.

\end{document}